\documentclass[letterpaper, 10 pt, conference]{ieeeconf} 
\usepackage{subcaption}
\usepackage{graphicx}
\usepackage{changepage}
\usepackage{booktabs}
\usepackage{afterpage}
\usepackage{balance}

\IEEEoverridecommandlockouts 

\usepackage{amsmath} 

\title{\LARGE \bf
TornadoDrone: Bio-inspired DRL-based Drone Landing on 6D Platform with  Wind Force Disturbances
}

\author{Robinroy Peter, Lavanya Ratnabala, Demetros Aschu, Aleksey Fedoseev, and Dzmitry Tsetserukou
\thanks{The authors are with the Intelligent Space Robotics Laboratory, Skolkovo Institute of Science and Technology, Bolshoy Boulevard 30, bld. 1, 121205, Moscow, Russia}
\thanks{
email: {(robinroy.peter, lavanya.ratnabala, demetros.aschu, aleksey.fedoseev, d.tsetserukou})@skoltech.ru}}

\begin{document}

\maketitle
\thispagestyle{empty}
\pagestyle{empty}


\begin{abstract}
Autonomous drone navigation faces a critical challenge in achieving accurate landings on dynamic platforms, especially under unpredictable conditions such as wind turbulence. Our research introduces TornadoDrone, a novel Deep Reinforcement Learning (DRL) model that adopts bio-inspired mechanisms to adapt to wind forces, mirroring the natural adaptability seen in birds. This model, unlike traditional approaches, derives its adaptability from indirect cues such as changes in position and velocity, rather than direct wind force measurements. TornadoDrone was rigorously trained in the gym-pybullet-drone simulator, which closely replicates the complexities of wind dynamics in the real world. Through extensive testing with Crazyflie 2.1 drones in both simulated and real windy conditions, TornadoDrone demonstrated a high performance in maintaining high-precision landing accuracy on moving platforms, surpassing conventional control methods such as PID controllers with Extended Kalman Filters. The study not only highlights the potential of DRL to tackle complex aerodynamic challenges but also paves the way for advanced autonomous systems that can adapt to environmental changes in real-time. The success of TornadoDrone signifies a leap forward in drone technology, particularly for critical applications such as surveillance and emergency response, where reliability and precision are paramount.
\\

\emph{Keywords — Autonomous Drone Landing, Deep Reinforcement Learning, Gym-PyBullet-Drone Simulation, Wind Disturbance, Moving Platforms, Bio-inspired Robots.}

\end{abstract}

\section{Introduction}

The field of autonomous drone navigation has experienced significant advancements in recent years, propelled by rapid progress in artificial intelligence and robotics. Unmanned Aerial Vehicles (UAVs) have become increasingly vital across a range of applications, including surveillance, delivery services, environmental monitoring, and emergency response \cite{b0}. A pivotal aspect of these applications is the drone's ability to perform precise and safe landings on moving platforms a task that continues to pose substantial challenges. This challenge is further intensified by complex wind forces \cite{b26}. 

Traditional control methods often struggle to dynamically adapt to these rapidly changing aerodynamic conditions \cite{b8}. 
\begin{figure}[ht]
\centering
\includegraphics[width=0.48\textwidth]{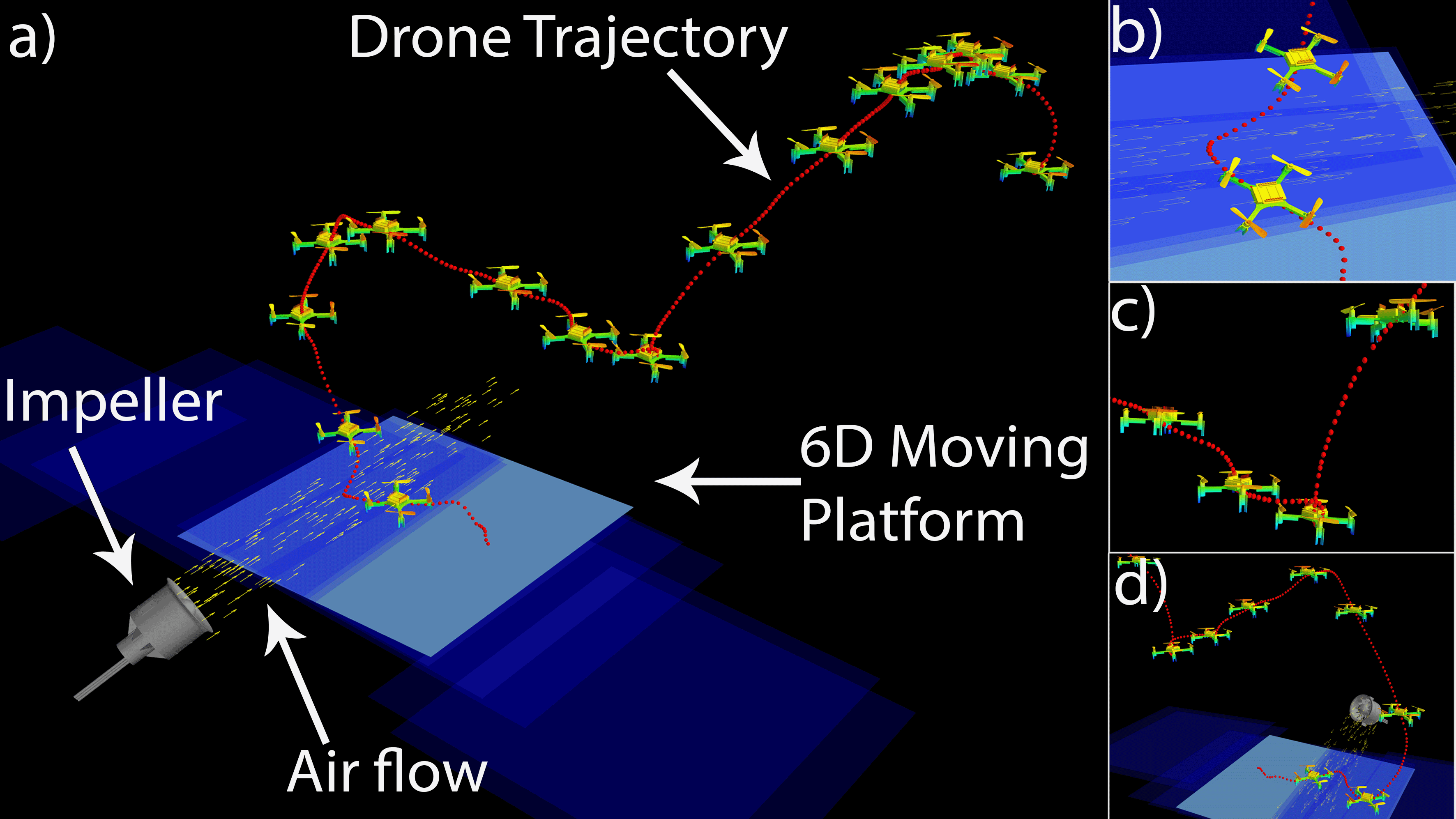}
%
 
\caption{Composite frame illustrating key phases of autonomous landing: (a) Complete trajectory overview, (b) Instant re-planning in response to external forces, (c) Sudden recovery behavior (d) Adaptation to sudden directional changes of the moving landing platform.}
 \label{fig:test}
 \end{figure}
Achieving accurate landings on a moving platform in the presence of the wind effect is not just a technical accomplishment, it represents a significant advancement in the operational capabilities of drones for real-world applications.


Deep Reinforcement Learning (DRL) has emerged as a promising approach for handling uncertainties. DRL involves training algorithms via a trial-and-error approach, enabling them to determine optimal actions in complex and unpredictable environments. This methodology is exceptionally suited for autonomous drone landing tasks, where drones must make real-time decisions based on immediate environmental feedback.
This research introduces the TornadoDrone agent (Fig.~\ref{fig:test}), utilizing DRL for effective landing and motion planning under unpredictable conditions, including sudden wind changes and moving platform velocities. We utilize the Vicon indoor localization system to test the agent in real-world conditions, it effectively guides the drone to land on a moving platform with uncertainties.

The agent's performance, trained in a simulated environment and tested on Crazyflie 2.1 drones, is benchmarked against a baseline PID controller with an Extended Kalman filter (EKF), demonstrating the model's adaptability. This work significantly advances drone autonomy and safety, potentially transforming their deployment in dynamic scenarios.

Our agent revolutionizes drone landing with bio-inspired learning, intuitively handling external forces like wind without exact force data. Its training transcends specific drone specs, ensuring broad adaptability through domain randomization. Our model deciphers indirect flight dynamics to seamlessly counter environmental challenges, mirroring birds' natural flight adaptability. This approach significantly enhances drone flexibility across diverse scenarios without custom modifications.

\section{Related Works}
The domain of autonomous UAV landing has seen considerable evolution, driven by advancements in vision-based techniques, DRL, and innovative landing strategies. This research spectrum extends from precision landings on static platforms to the dynamic challenges of moving platforms, with an emphasis on environmental adaptability. 

A study on autonomous land on a moving vehicle using a visual servoing controller that processes velocity commands directly in image space \cite{b21}. A vision-based drone swarm docking system \cite{c28} designed to enable robust landing on a moving platform. Vision-based approaches, as delineated by authors of \cite{b8}, \cite{b21}, and \cite{b1}, have been effectively merged with DRL to enhance UAV landing capabilities. These methods demonstrate the potential of integrating real-time visual inputs with a DRL for precise landings, and their performance can be affected by environmental factors such as lighting and weather conditions. 

DRL is a promising approach across various research domains within UAV studies, particularly in areas like drone racing \cite{b22}, \cite{b14}, perching \cite{b12}, \cite{b29}, path planning \cite{b23}, and in the area of drone landing.

Earlier researchers often focused on static platform landings. Gazebo-based RL framework was developed by the author of \cite{b6} for drone landing. Authors of \cite{b1}, \cite{b2}, \cite{b3}, \cite{b7} and \cite{b27} showcasing significant advancements in this area. These works, however, primarily concentrate on the precision aspect without extensively tackling the unpredictable dynamics associated with moving platforms. 
The challenges of landing on moving platforms have prompted the development of adaptive algorithms, as explored in the works in \cite{b4} and \cite{b5}. However, these investigations often overlook the impact of external factors like wind, a critical element in UAV landing dynamics. Recent studies such as those by authors of \cite{b26}, have addressed the turbulence effects on UAV aerodynamics, yet a comprehensive treatment that intertwines wind forces with moving platform dynamics is notably absent.

The study in \cite{b28} discusses the concept of bio-inspired intelligent microdrones that can perform complex tasks autonomously using simple sensors with low computing power. Some studies show the application of bio-inspired behavior perching in UAV \cite{b12}, \cite{b29}. Authors of \cite{b30} discussed how insects use spatial knowledge about the environment to do the navigation.

Our research introduces a bio-inspired DRL framework that addresses the intricacies of autonomous landing amid environmental disturbances, particularly wind. Drawing inspiration from avian adaptability, our approach endows UAVs with the ability to intuitively navigate wind disturbances without explicit force measurements. This adaptability, coupled with our framework's model-agnostic training in environments like the gym-pybullet-drone simulator \cite{b15} which accounts for factors like downwash and drag \cite{b16}—marks a significant stride in UAV landing technology.

Moreover, our work extends the discourse on UAV adaptability in dynamic conditions, integrating insights from research on various UAV tasks and environmental interactions \cite{b7}, \cite{b12}, \cite{b14}, \cite{b30} and \cite{b31}. By addressing the limitations of existing methodologies and proposing a solution that considers environmental dynamics and external wind forces, our study contributes to the advancement of UAV operational safety, reliability, and efficiency in real-world conditions.

\section{Methodology}

This section outlines the approach employed to develop and validate the TornadoDrone agent, focusing on autonomous drone landing on dynamically moving platforms under varying environmental conditions using the DRL framework.

\subsection{Simulation Environment Setup}
\begin{figure}[h]
 \centering
 \includegraphics[width=0.3\textwidth]{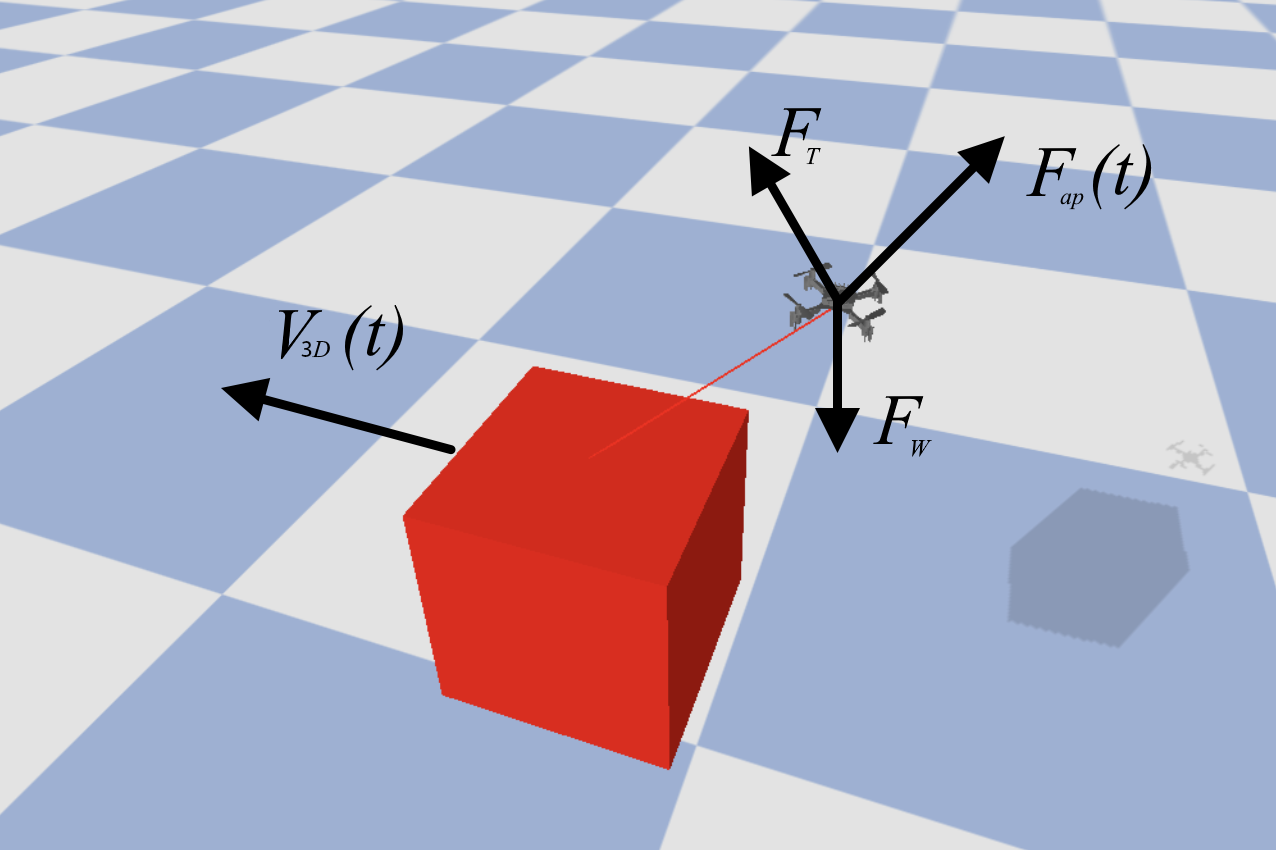}
 \caption{Simulation setup in gym-pybullet environment.}
 \label{fig:land}
\end{figure}

Our simulation environment is developed using the Gym framework and PyBullet physics, featuring configurations that emulate the Crazyflie 2.x drone for realistic aerodynamic simulations, using gym-pybullet-drones \cite{b15}. Central to our setup is a 0.5 m cubical platform, depicted in Fig.~\ref{fig:land}, which moves in the XYZ space with velocities ranging from -0.46 to 0.46 m/s, thus introducing a dynamic challenge for precision landings.

The drone's parameters include linear velocities from -3 to 3 m/s in XY and -2 to 2 m/s in Z, and rotation angles from $-\pi$ to $\pi$ radians, necessitating adaptive flight strategies for effective landings. To enhance the TornadoDrone agent's training and adaptability to external disturbances, we introduced a probabilistic framework for applying random external forces, as in:

\begin{equation}
F =
\begin{cases}
F_{ap}(t) & \text{if } p(e) < 0.2 \\
0 & \text{if } p(e) \geq 0.2
\end{cases} ,
\end{equation}

\begin{equation}
F_{ap}(t) =
\begin{cases}
\text{sgn}(f(t,\xi)) \times |\text{f}(t, \xi)| & \text{if } p(s) < 0.2 \\
0 & \text{if } p(s) \geq 0.2 
\end{cases} ,
\end{equation}
where $p(e)$ is the probability at which the force will be applied during the episode, $p(s)$ is the probability that the force will be applied at the current step of the “windy" episode, $F$ is the vector of an external force based on the binary indicator $f_{i}$. The force direction is selected randomly with $x$, $y$, and $z$ components in the world coordinate frame being in the range of -0.005 to 0.005, simulating realistic environmental disturbances like wind. This method aims to increase the agent's resilience and performance under varied and unpredictable conditions typical in real-world operational scenarios.

\subsection{Deep Reinforcement Learning Framework}

\subsubsection{TornadoDrone Agent Architecture}
The TornadoDrone agent employs a neural network tailored for drone landing, utilizing observations as follows: 
\begin{equation}
 \vec{o}_t = \left[ \vec{\theta}, \vec{v}, \vec{\omega}, \vec{d}, \Delta \vec{v} \right],
\end{equation}
where $\vec{\theta}$ is the attitude of the drone (roll, pitch, yaw), $\vec{v}$ is the linear velocity, $\vec{\omega}$ is the angular velocity, $\vec{d}$ are the relative landing pad positions, and $\Delta \vec{v}$ are the relative velocities of the landing pad. These inputs are first clipped and then normalized to a range of -1 to 1, ensuring optimal neural network performance.

\begin{figure}[h]
 \centering
 \includegraphics[width=0.4\textwidth]{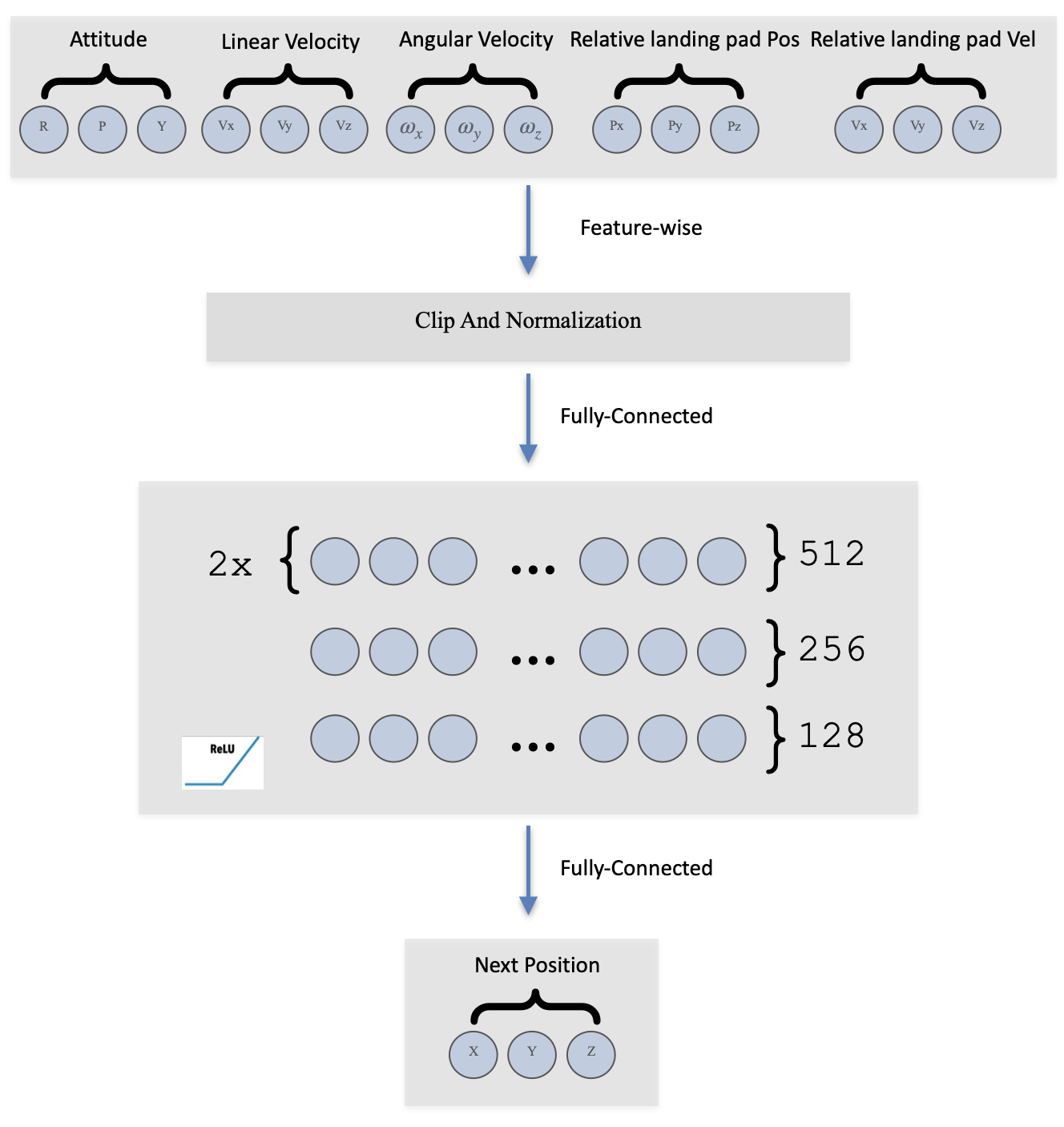}
 \caption{Architecture of the DRL model illustrating inputs, hidden layers, and action mechanisms.}
 \label{fig:rl}
\end{figure}

Fig.~\ref{fig:rl} shows the neural network architecture behind our DRL approach.
\begin{equation}
\resizebox{0.9\linewidth}{!}{$\text{ReLU}(\text{FC}{512 \times 2}(\hat{\vec{o}}t)) \rightarrow \text{ReLU}(\text{FC}{256}) \rightarrow \text{ReLU}(\text{FC}{128})$} 
\end{equation}
where FC are the fully-connected layers with ReLU activation functions, arranged in dimensions of 512x2, 256, and 128. This setup processes the standardized inputs to determine the drone's precise adjustments for landing. The output layer, with three neurons, generates PID control signals dictating position changes in the range of -1, 0, and 1. 
\begin{equation}
\Delta \vec{p}_t = 0.1 \times \vec{c}_t , 
\end{equation}
where $\vec{c}_t$ is the control signal for position change $\Delta \vec{p}_t$. These adjustments are applied to the current drone pose, with a 0.1 factor, guiding the drone towards an accurate landing.

\subsubsection{Reward Function}
The agent's reward function is crafted to enhance precision and adaptability in landing. It is structured as follows:

\begin{equation}
\resizebox{0.9\linewidth}{!}{$
\begin{aligned}
\text{Reward} = 
 \begin{cases}
 \tanh(\gamma), & \text{if } d_{target} > 2 \\
 \tanh(\alpha \times (d_{target} - R)), & \text{if } d_{target} \subseteq (0.1, 2) \\
 \tanh(-U - \beta + \Delta), & \text{if } d_{target} < 0.1\\
 \tanh(-U + \Delta), & \text{Otherwise}
 \end{cases}
\end{aligned}$}
\end{equation}
where $\gamma$ is the penalty reward for moving far away from the target, $d_{target}$ is the distance between the drone and the target landing pad, $\alpha$ is the reward scaling factor for proximity to the target, $R$ is the current distance to the target. $U = U_{attractive} + U_{repulsive}$ combines attractive and repulsive potentials. $\beta$ adjusts for edge proximity penalties and below the landing pad altitude. $\Delta$ discourages excessive speed allowing descending relative velocity while approaching the landing pad.


The attractive and repulsive potentials are defined as:
\begin{equation}
\resizebox{0.9\linewidth}{!}{$
U_{repulsive} = 
\begin{cases} 
 \frac{1}{2} \times \eta \times \left( \frac{1}{\sigma} - \frac{1}{Q_{max}} \right)^2, & \text{if } \sigma < Q_{max} \\
 0, & \text{Otherwise}
\end{cases}$}
\end{equation}

\begin{equation}
U_{attractive} = \frac{1}{2} \times \zeta \cdot R^2
\end{equation}
where $\eta$ is the strength of the repulsive potential, $\sigma$ is the distance to the nearest obstacle, $Q_{max}$ is the maximum effective distance of the repulsive potential, $\zeta$ is the strength of the attractive potential, $R$ is the current distance to the target. This reward function dynamically balances the TornadoDrone agent's objectives, guiding it toward successful landings while avoiding hazards and ensuring smooth descent trajectories.
\begin{figure}[h]
 \centering
 \begin{minipage}{0.45\textwidth}
 \centering
 \includegraphics[width=\textwidth]{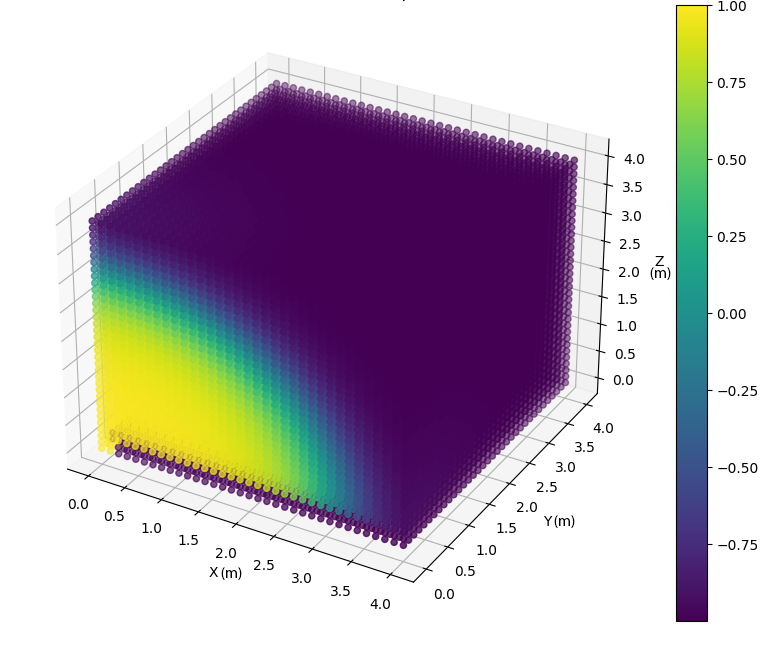}
 \caption{Origin view of TornadoDrone's reward function emphasizing safety and behavior.}
 \label{fig:reward1}
 \end{minipage}\hfill 
\end{figure}

The reward function of the TornadoDrone agent, integral to our methodology, is constructed using a potential field approach and is depicted in 3D space for comprehensive visualization. Currently, the function employs an attractive potential field to guide the drone towards its landing target, with a reward gradient that enhances precision by offering higher rewards closer to the target zone, as illustrated in Fig.~\ref{fig:reward1}. This gradient is apparent from the origin view, with the color transition from yellow to dark purple indicating the shift from optimal to less desirable states.

In addition to the attractive potential field, a crucial safety mechanism is embedded within the reward structure, which penalizes the drone from operating below a predefined altitude relative to the landing pad. This safety reward ensures that the drone maintains a safe approach trajectory and does not fly at an altitude that would be considered hazardous or below the landing platform's level.

Looking ahead, the reward function is poised for expansion to include a repulsive potential field. This future development aims to further sophisticate the agent's navigational capabilities by introducing negative rewards for approaching obstacles, thereby preventing collisions and reinforcing safe flight paths in complex environments.
\subsubsection{Training Protocol}

The TornadoDrone agent training protocol is meticulously structured to ensure an effective learning progression. Using the stable-baselines3 Twin Delayed DDPG (TD3) algorithm, the agent undergoes a rigorous training regimen designed for complex and continuous control tasks. The choice of TD3 was motivated by its demonstrated ability to converge more rapidly compared to alternative algorithms, particularly in environments with continuous and complex behaviors.

The policy employed is the MlpPolicy with an initial learning rate of 0.0001. The agent was initially trained in over 5 million steps, with each episode capped at 20 seconds to allow the agent to acquire the main behavioral patterns necessary for landing. To further refine the agent's capabilities, including additional safety maneuvers and adaptability skills, the model underwent retraining up to 35 million steps. This extended training involved multiple iterations of fine-tuning in both simulated environments and real-world testing scenarios, enhancing performance in dynamic 3D spaces where the landing pad presents complex patterns and sudden directional changes. An overview of the parameter configuration is shown in  Table ~\ref{table:training_protocol}.

\begin{table}[h]
\centering
\caption{\centering{Training Parameters and DRL Algorithm Configurations.}}
\begin{tabular}{l l}
\hline
\textbf{Parameter} & \textbf{Value} \\
\hline
Algorithm & Twin Delayed DDPG (TD3) \\
Policy & MlpPolicy \\
Learning Rate & 0.0001 \\
Initial Training Steps & 5 million \\
Extended Training Steps & Up to 35 million \\
Episode Duration & 20 s \\
Buffer Size & 1,000,000 (1e6) \\
Batch Size & 100 \\
Activation Function & ReLU \\
Feature Extractor & FlattenExtractor \\
Optimizer & Adam \\
\hline
\end{tabular}
\label{table:training_protocol}
\end{table}

The training leverages a buffer size of 1,000,000 (1e6), with learning commencing after 100 steps and a batch size of 100. The MlpPolicy parameters are set to use ReLU activation functions, a FlattenExtractor for feature extraction, and image normalization is enabled. The optimizer of choice is Adam, known for its efficiency in handling sparse gradients on noisy problems.

\begin{figure}[h]
 \centering
 \begin{minipage}{0.48\textwidth}
 \centering
 \includegraphics[width=\textwidth]{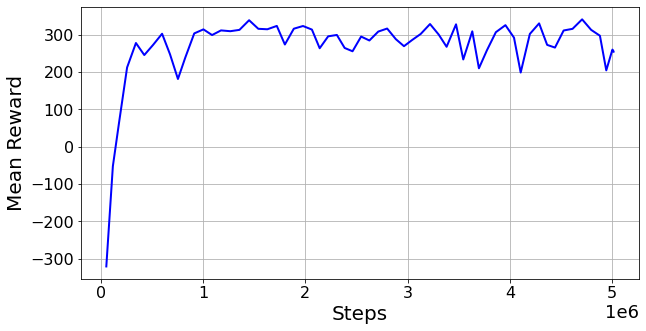}
 \caption{Mean reward vs training steps, showcasing learning progress.}
 \label{fig:rewardstep}
 \end{minipage}\hfill
\\
 \begin{minipage}{0.48\textwidth}
 \centering
 \includegraphics[width=\textwidth]{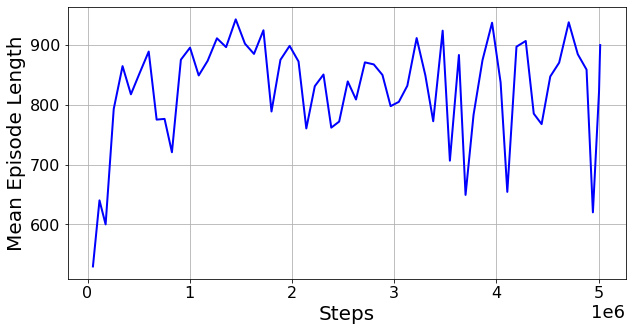}
 \caption{Mean episode length vs training steps, indicating agent endurance.}
 \label{fig:epstep}
 \end{minipage}
\end{figure}

This training protocol culminates in agents navigating in 3D spaces with dynamic landing platforms, demonstrating quick adaptation to unforeseen environmental changes and complex landing trajectories.
Fig.~\ref{fig:rewardstep} and Fig.~\ref{fig:epstep} capture the agent's learning progress, with an increasing mean reward and episode length over training steps showcasing the agent's enhanced reward optimization and sustained performance throughout the learning phase.

\subsection{Real-World Validation Setup}
\subsubsection{Indoor Localization System (Vicon)}

For real-world testing, we employed a Vicon motion capture system to provide high-precision localization of both drones and platforms. This system delivers positional data at a rate of 100Hz, which is critical for extracting accurate observations necessary for the TornadoDrone agent's operation. The observations are then fed into the agent to inform its decision-making process. The Vicon system's VRPN (Virtual-Reality Peripheral Network) positioning type ensures a robust and precise tracking capability, vital for the successful deployment and testing of our autonomous navigation algorithms in a controlled indoor environment.

\subsubsection{Crazyflie Drones}

The empirical tests were conducted using Crazyflie2.1 drones, which are equipped with onboard default PID controllers for low-level flight control. The system was integrated with ROS2, which facilitated the issuance of high-level position commands. Communication with the drones was achieved through a Crazyradio 2.4 GHz RF transmitter, operating at a frequency of 100Hz to ensure real-time responsiveness and precise maneuvering during flight tests.

\subsection{Baseline Comparison}
\subsubsection{Crazyflie On-board PID Controller with EKF}

The baseline for our comparative analysis incorporates a PID control system enhanced with an EKF for precise tracking of moving landing platforms. The EKF implementation is tailored to predict and update the platform's position and velocity, ensuring accurate tracking under dynamic conditions. The core of the EKF is defined and initialized with the platform's initial state, covariance, and the variances associated with the process and measurements.

The state transition matrix $\mathbf{A}$ and observation matrix $\mathbf{H}$ are constructed as follows:
\[
\mathbf{A} = \begin{bmatrix}
 1 & 0 & 0 & 1 & 0 & 0 \\
 0 & 1 & 0 & 0 & 1 & 0 \\
 0 & 0 & 1 & 0 & 0 & 1 \\
 0 & 0 & 0 & 1 & 0 & 0 \\
 0 & 0 & 0 & 0 & 1 & 0 \\
 0 & 0 & 0 & 0 & 0 & 1
\end{bmatrix}, \quad
\mathbf{H} = \begin{bmatrix}
 1 & 0 & 0 & 0 & 0 & 0 \\
 0 & 1 & 0 & 0 & 0 & 0 \\
 0 & 0 & 1 & 0 & 0 & 0
\end{bmatrix}
\]

The `predict' method advances the state estimation based on the motion model, while the `update' method refines this estimation with incoming measurements, employing the Kalman Gain to minimize the estimation error. This EKF framework serves as a robust baseline, facilitating a comprehensive evaluation of the TornadoDrone agent's performance in tracking and landing on moving platforms.

\section{Experiments}
Our experiments are designed to explore key aspects of autonomous drone landing: (i) Comparing the TornadoDrone agent's landing accuracy and consistency against traditional control methods on moving platforms. (ii) Assessing the TornadoDrone agent's resilience to environmental disturbances and dynamic platform behaviors. (iii) Evaluating the agent's versatility across varied and complex landing scenarios. (iv) Validating the agent's simulation-trained strategies in real-world settings.

\subsection{Experimental Design}
Our real-world testing framework was meticulously designed to validate the TornadoDrone agent under various conditions. Utilizing a UR10 robotic arm, we mounted a $0.5 \times 0.5 \times 0.003$ meter acrylic landing pad on its TCP, ensuring precise and controlled movements. To simulate air disturbances, an impeller (Fms 64Mm Ducted Fan System 11-Blade W/2840-Kv3900 Motor FMSDF004) powered by a 12V battery through an Arduino Uno was embedded in the UR10 robotic arm, positioned 0.3 meters from the center of the landing pad, directing airflow towards the pad's center. This setup aims to create realistic wind disturbances affecting the drone during landing maneuvers.

The experimental setup was divided into six distinct scenarios to comprehensively evaluate landing performance, with 10 - 15 test cases per scenario, overall 120 test cases:

\begin{enumerate}
 \item {Static Point Landing (SPL)}: Testing the agent's ability to land on a stationary platform.
 \item {Linear Moving Point Landing (LMPL)}: Assessing landings on a platform moving linearly with sudden directional changes.
 \item {Curved Moving Point Landing (CMPL)}: Evaluating landings on a platform following a curved trajectory with directional shifts.
 \item {Complex Trajectory Landing (CTL)}: The TornadoDrone agent's adaptability is further tested through challenging landings on dynamically moving platforms in three-dimensional space, amidst wind disturbances generated by impellers mounted on the landing pads.
 \item {Static Point Landing with Wind Disturbance (SPL-WD)}: This scenario introduces additional tests for landing on a stationary platform under two conditions of wind disturbance, simulated by impeller speeds of 4500 rpm and 8500 rpm.
 \item {Linear Moving Point Landing with Wind Disturbance (LMPL-WD)}: Here, the agent is tested on a linearly moving platform under two wind disturbance conditions, with impeller speeds set to 4500 rpm and 8500 rpm, to assess its adaptability and control under increased environmental complexity.
\end{enumerate}

\begin{figure}
\centering
\begin{subfigure}[b]{0.15\textwidth}
\includegraphics[width=\textwidth]{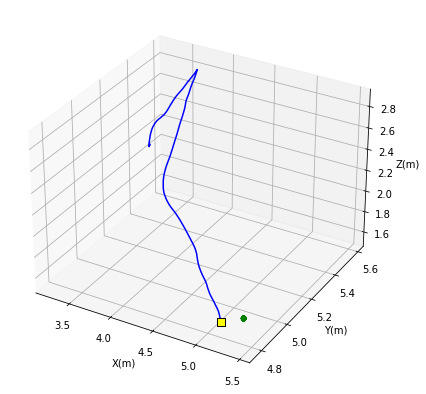}
\caption{}
\end{subfigure}
\begin{subfigure}[b]{0.15\textwidth}
\includegraphics[width=\textwidth]{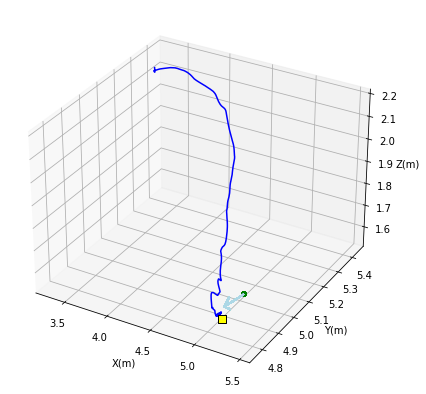}
\caption{}
\end{subfigure}
\begin{subfigure}[b]{0.15\textwidth}
\includegraphics[width=\textwidth]{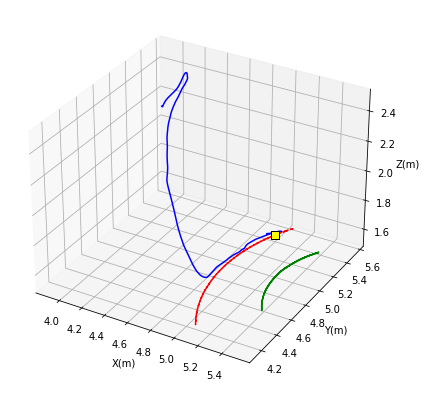}
\caption{}
\end{subfigure}
\\
 \begin{subfigure}[b]{0.15\textwidth}
 \includegraphics[width=\textwidth]{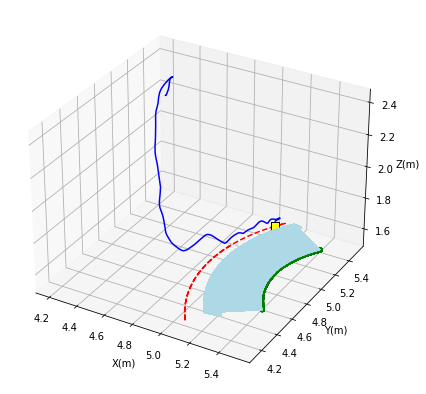}
 \caption{}
 \end{subfigure}
 \begin{subfigure}[b]{0.15\textwidth}
 \includegraphics[width=\textwidth]{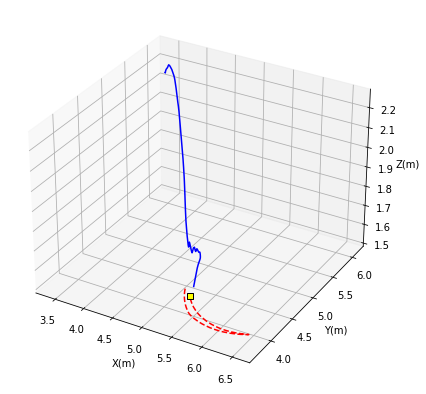}
 \caption{}
 \end{subfigure}
 \begin{subfigure}[b]{0.15\textwidth}
 \includegraphics[width=\textwidth]{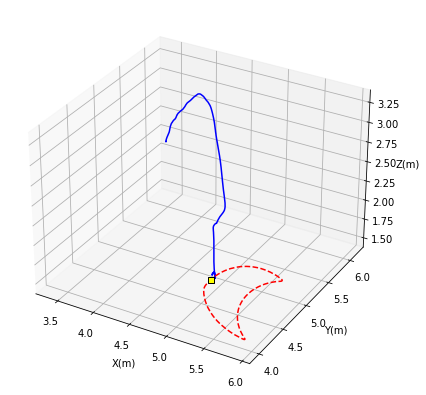}
 \caption{}
 \end{subfigure}

\caption{Trajectories of the drone (blue line) and landing pad (red line) impeller position (green) in (a) SPL without wind, (b) SPL with the wind (c) LMPL without wind (d) LMPL with the wind (e) CMPL without wind (f) CTL without wind}
 \label{fig:traj1}
 \end{figure}

Comparing our agent's performance against a baseline established using an EKF with the onboard Crazyflie PID controller. To further test the agent's adaptability, scenarios 5 and 6 were also conducted with air disturbances generated by the impeller, exclusively for the TornadoDrone agent.

\subsection{Performance Metrics}
To evaluate the algorithm's effectiveness against the baseline controller, we employed a comprehensive set of performance metrics:

 \subsubsection{Landing Success Rate}
 We evaluate the agent's ability to successfully land on dynamic platforms under various conditions, including those with wind disturbances. This metric reflects the agent's reliability and consistency throughout the experiments.

\begin{table}[h!]
\centering
\caption{\centering{Landing Success Rates: TornadoDrone Agent vs. EKF with PID Controller}}
\begin{tabular}{|l|l|l|}
\hline
\textbf{Test Case} & \textbf{TornadoDrone Agent} & \textbf{EKF with PID} \\ \hline
SPL & 100\% & 80\% \\ \hline
LMPL & 100\% & 20\% \\ \hline
CMPL & 100\% & 40\% \\ \hline
CTL & 60\% & 10\% \\ \hline
SPL-WD (4500 rpm) & 100\% & N/A \\ \hline
SPL-WD (8500 rpm) & 100\% & N/A \\ \hline
LMPL-WD (4500 rpm) & 91.67\% & N/A \\ \hline
LMPL-WD (8500 rpm) & 78.57\% & N/A \\ \hline
\end{tabular}
\label{table:landing_success_rate}
\end{table}

As depicted in Table \ref{table:landing_success_rate}, the TornadoDrone agent consistently achieves high landing success rates, markedly surpassing the performance of the traditional EKF-PID control system across various scenarios. This superiority is evident not only in standard conditions but also in tests involving wind disturbances, underlining the agent's robust adaptability and precise control capabilities.

\subsubsection{Landing Precision} This metric assesses how accurately the drone can reach a designated target on the moving platform. Precision is measured by the average distance from the target across several landing attempts, with the standard deviation indicating the consistency of these landings.

\begin{table}[h]
\caption{\centering{Landing Precision Comparison: TornadoDrone Agent vs EKF with PID}}
\centering
\begin{tabular}{|l|c|c|c|c|} 
 \hline
\textbf{Test Case} & \multicolumn{2}{c|}{\textbf{TornadoDrone Agent}} & \multicolumn{2}{c|}{\textbf{EKF with PID}} \\ 
 \hline
 & Mean (cm) & STD (cm) & Mean (cm) & STD (cm) \\ 
 \hline
SPL & 3.72 & 0.15 & 10.32 & 2.85 \\ 
 \hline
LMPL & 4.91 & 1.62 & 9.35 & 4.79 \\ 
 \hline
CMPL & 7.14 & 1.82 & 10.26 & 5.60 \\ 
 \hline
CTL & 10.41 & 4.06 & 15.28 & 3.91 \\ 
 \hline
\end{tabular}
\label{table:landing_precision}
\end{table}
\begin{table}[h!]
\caption{\centering{Landing Precision with Wind Disturbance}}
\centering
\begin{tabular}{|l|l|l|l|}
\hline
\textbf{Test Case (rpm)} & \textbf{Min (cm)} & \textbf{Mean (cm)} & \textbf{STD (cm)} \\ 
\hline
SPL-WD (4500) & 2.38 & 3.96 & 1.58 \\
\hline
SPL-WD (8500) & 3.48 & 6.31 & 1.99 \\ 
\hline
LMPL-WD (4500) & 2.65 & 5.88 & 5.18 \\ 
\hline
LMPL-WD (8500) & 3.26 & 9.22 & 9.47 \\ 
\hline
\end{tabular}
\label{table:airforce}
\end{table}
Tables \ref{table:landing_precision} and \ref{table:airforce} detail the TornadoDrone agent's landing precision compared to the EKF with PID controller, covering both standard and wind-disturbance-enhanced scenarios. The TornadoDrone agent showcases significantly better precision in all tested conditions, emphasizing its robustness and capability to navigate and land accurately in complex, dynamic environments.
\afterpage{
\begin{figure*}[t]
 \centering
 
\includegraphics[width=0.9\textwidth]{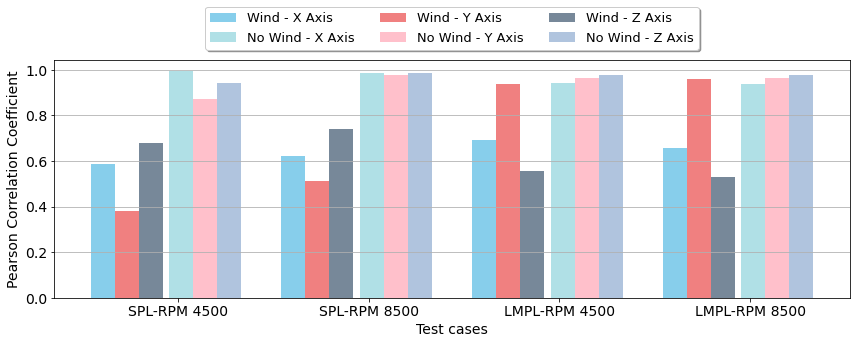}
 \caption{Pearson averaged correlation for drone position and model predicted position}
 \label{fig:pearson}
\end{figure*}

}
 
 \subsubsection{Complexity of Scenarios and Recovery from Perturbations} The agent's performance is tested against a spectrum of complex situations, including unpredictable target movements and challenging environmental conditions, as well as its capacity to stabilize and land following disturbances such as wind gusts or abrupt target motion changes, to ascertain its versatility, real-world applicability, adaptability, and resilience. A summary of the correlation between drone velocity vs landing pad velocity statistics and comments are provided in Table \ref{table:summary}.
 \begin{figure}[h]
 \centering
 \includegraphics[width=0.45\textwidth]{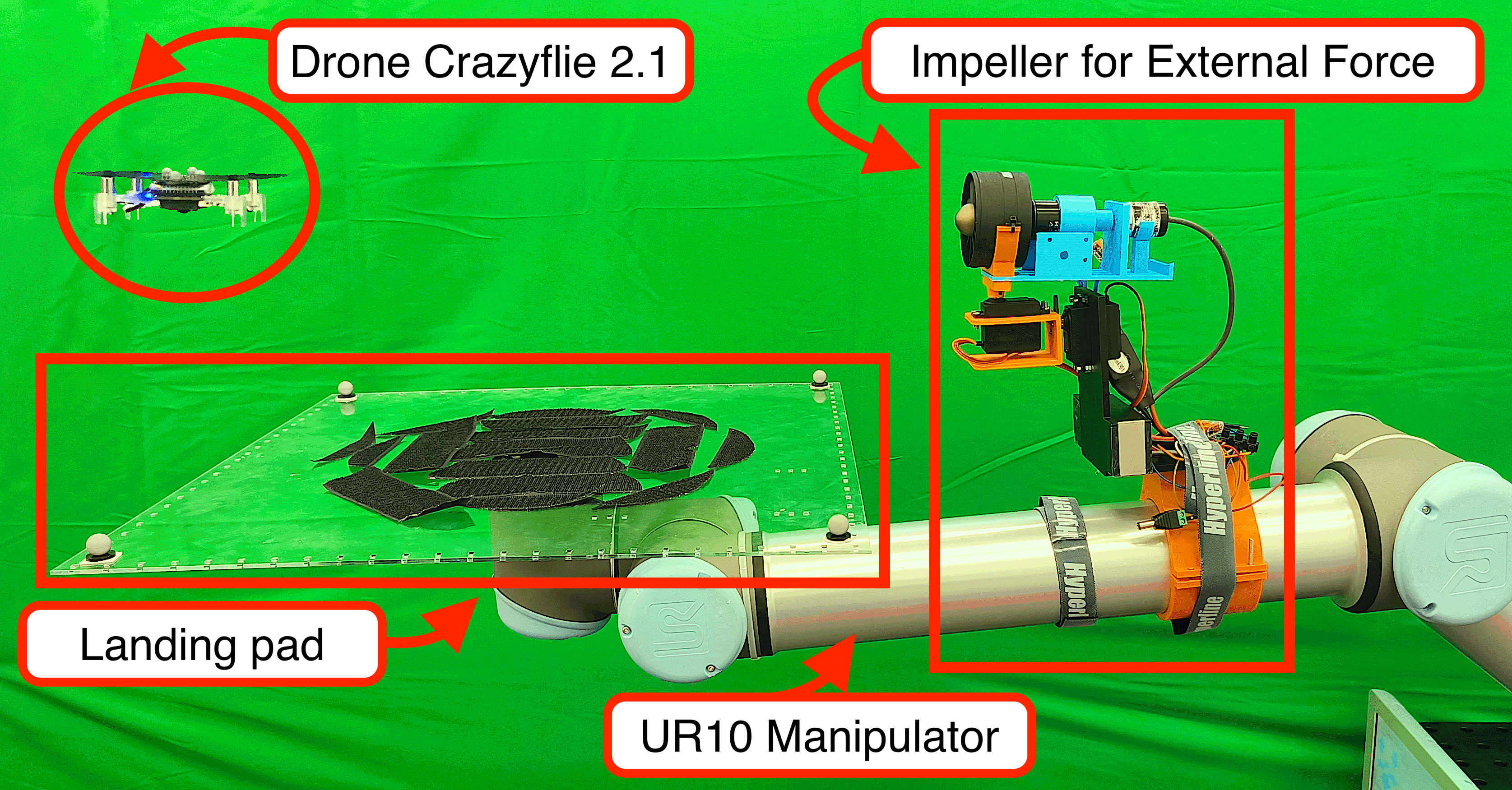}
 \caption{Experimental setup with drone landing on a moving platform in the presence of external force.}
 \label{fig:reallanding}
\end{figure}

 \begin{table}[h!]
 \caption{\centering{Summary for Complexity of Scenarios and Recovery from Perturbations}}
 \centering
 \begin{tabular}{|l|c|c|c|c|c|}
 \hline
\textbf{ Correlation} & \textbf{SPL} & \textbf{LMPL} & \textbf{CMPL} & \textbf{CTL} \\ \hline
 Mean & 0.1012 & 0.5822 & 0.5028&0.2340 \\ \hline
 Median & 0.1020 & 0.6055 &0.4947 &0.2153 \\ \hline
 STD & 0.0674 & 0.1174 &0.2320 & -0.2399 \\ \hline
 Min & -0.0067 & 0.3561 & -0.0468 & -0.1106\\ \hline
 Max & 0.1829 & 0.7407 & 0.7703 & 0.5369 \\ \hline
 \end{tabular}
 \label{table:summary}
\end{table}
The real-world experimental setup with a drone landing on a moving
platform in the presence of external force is demonstrated in Fig.~\ref{fig:reallanding}.
The results of the experiment display a correlation between drone and landing pad velocities across diverse test cases assessing the TornadoDrone agent's adaptability.

\begin{figure}[ht]
 \centering
 \includegraphics[width=0.4\textwidth]{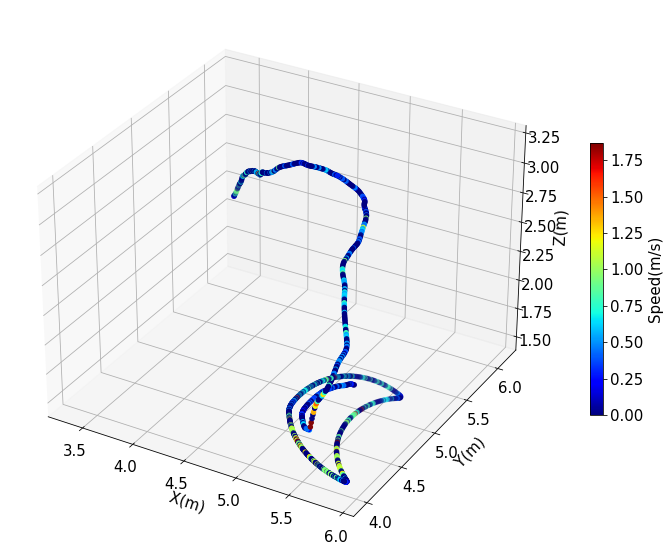}
 \caption{Velocity changes of drones and moving pad explaining the complexity}
 \label{fig:velocity}
\end{figure}

Notably, higher mean correlations suggest better synchronization. Strong adaptability is evident in linear moving point landings, while variability in correlation coefficients reflects resilience levels. Standard deviation and correlation ranges provide insights into consistency and robustness. These metrics offer quantitative assessments of the TornadoDrone agent's performance in varied environments and perturbations.
The velocity changes and their complexities are illustrated in Fig.~\ref{fig:velocity}. 

The velocity analysis reveals that the highest velocity snap of 1.75 m/s happens during the final stage of the landing where the agent adapts to the ground effect and external force simultaneously. However, the change in trajectory at this point is compensated by the DRL agent, showcasing a successful landing performance. The adaptability of the developed agent to various scenarios is illustrated through drone landing trajectories in Fig.~\ref{fig:traj1}, covering fixed point, fixed point with wind, linear, linear with wind, curve, and complex 3D motions.

\subsubsection{Bio-Inspired Wind Force Recognition} A performance metric we introduced is the evaluation of the TornadoDrone agent's ability to recognize and adapt to external wind forces in a bio-inspired manner, akin to how birds perceive and react to environmental changes through their internal states. To quantify this capability, we measured the Pearson correlation between the drone's actual position and the positions predicted by our DRL model throughout the experiment. This analysis was segmented into wind-affected and no-wind areas, providing insights into the model's responsiveness to wind disturbances across the x, y, and z axes.

The correlation values in wind-affected areas versus no-wind areas offer a direct measure of the model's sensitivity to wind-induced positional deviations. High correlation values in no-wind areas indicate accurate positional prediction under stable conditions. In contrast, the correlation values in wind-affected areas reveal how well the model infers and compensates for the wind's impact, mirroring a bird's instinctive adjustments to maintain its flight path.

Analysis of the correlation data, as illustrated in Fig.~\ref{fig:pearson}, underscores the TornadoDrone agent's proficiency in discerning and counteracting wind forces through its internal state adjustments. For instance, in scenarios such as SPL-WD and LMPL-WD, the distinct correlation patterns in the presence versus absence of wind elucidate the model's dynamic adaptability. This bio-inspired recognition and response mechanism not only validates the agent's effectiveness in navigating wind disturbances but also highlights its overall success in achieving its intended landing objectives. Through this, we demonstrate the TornadoDrone agent's sophisticated capability to emulate natural biological processes in sensing and adapting to environmental challenges, marking a significant stride in bio-inspired autonomous drone navigation.

\section{Conclusion and Future Work}
In this study, TornadoDrone demonstrated exceptional capabilities in autonomous drone landing, surpassing traditional control methods such as EKF with PID controllers in various performance metrics. Our agent achieved perfect success rates in static and linear moving platforms while considering wind disturbances and exhibited commendable performance in complex trajectory landing (CTL) scenarios with a 60\% success rate but our baseline struggled to get only a 10\% success rate. It shows how it adapts to dynamically moving landing pads. Moreover, the TornadoDrone has shown superior landing precision, achieving mean distances in complex trajectory landing scenarios of 10.41 cm and mean distances as low as 3.92 cm in SPL scenarios. It maintains high precision under wind-disturbed conditions with a remarkable accuracy of 3.96 cm in SPL-WD at 4500 rpm. This level of precision significantly outperforms the traditional EKF with the PID controller setup, which had a mean precision of 10.32 cm in SPL scenarios without wind disturbance. On a linearly moving platform with the presence of wind disturbance 8500 rpm model was performed as a best case with a 3.2 cm distance. The agent's adeptness in synchronizing with moving platforms was particularly evident in LMPL scenarios, where a notable mean correlation of 0.5822 was observed between the drone's velocity and the landing pad's velocity, highlighting recovery from perturbations. Additionally “Bio-Inspired Wind Force Recognition" performance metric has further validated the TornadoDrone's capability to adapt to external wind forces in a manner reminiscent of natural avian responses. We achieved the Pearson correlation coefficient for model predicted and actual drone action x, y, and z values of more than 0.9, and in the presence of wind disturbance achieved more than 0.5, showcasing how our model strongly reacts to it in the real world. From these, we can confirm the model adaptations to wind forces.
\balance

These results underscore the potential of TornadoDrone in enhancing the precision and reliability of drone landings, making it an invaluable tool for critical applications such as emergency response and logistics. The agent's bio-inspired approach to recognizing and adapting to environmental changes represents a significant stride towards more intelligent and adaptable autonomous drone systems. Future work will focus on further refining these capabilities, exploring unsupervised and meta-learning techniques to enable real-time adaptation to unforeseen environmental challenges, and broadening the scope of autonomous drone applications in complex, dynamically changing environments.






\addtolength{\textheight}{-12cm}
\end{document}